# From Classification to Generation: An Open-Ended Paradigm for Adverse Drug Reaction Prediction Based on Graph-Motif Feature Fusion

Yuyan Pi[1], Min Jin[1*], Wentao Xie[1], Xinhua Liu[1]

1 College of Computer Science and Electronic Engineering, Hunan University, Changsha, 410082, Hunan, China

*Corresponding authors: jinmin@hnu.edu.cn;

Abstract：

　　Computational biology offers immense potential for reducing the high costs and protracted cycles of new drug development through adverse drug reaction (ADR) prediction. However, current methods remain impeded by drug data scarcity-induced cold-start challenge, closed label sets, and inadequate modeling of label dependencies. Here we propose an open-ended ADR prediction paradigm based on Graph-Motif feature fusion and Multi-Label Generation (GM-MLG). Leveraging molecular structure as an intrinsic and inherent feature, GM-MLG constructs a dual-graph representation architecture spanning the atomic level, the local molecular level (utilizing fine-grained motifs dynamically extracted via the BRICS algorithm combined with additional fragmentation rules), and the global molecular level. Uniquely, GM-MLG pioneers transforming ADR prediction from multi-label classification into Transformer Decoder-based multi-label generation. By treating ADR labels as discrete token sequences, it employs positional embeddings to explicitly capture dependencies and co-occurrence relationships within large-scale label spaces, generating predictions via autoregressive decoding to dynamically expand the prediction space. Experiments demonstrate GM-MLG achieves up to 38% improvement and an average gain of 20%, expanding the prediction space from 200 to over 10,000 types. Furthermore, it elucidates non-linear structure-activity relationships between ADRs and motifs via retrosynthetic motif analysis, providing interpretable and innovative support for systematic risk reduction in drug safety.

1. Introduction

Adverse Drug Reactions (ADRs) refer to unexpected and noxious responses that appear in patients under the premise that the drug quality is qualified and the recommended usage and dosage are strictly followed. As a major global public health issue, despite the strictness of clinical trials and regulatory review processes, ADRs have become one of the leading factors causing high mortality and morbidity rates ranking after cancer and cardiovascular diseases. New drug research and development (R&D) and market launch are confronted with inherent challenges such as high investment, protracted cycles, and high risks, where approximately 90% of candidate drugs fail to pass Phase I clinical trials, a considerable part of which is attributed to the fact that drug entities with novel structures may induce unforeseen adverse reactions. Given the medication risks and enormous economic losses brought by ADRs, utilizing computational methods to improve the accuracy of ADR prediction and reduce unknown risks has emerged as a pivotal research focus.

Structure-Activity Relationship (SAR) and Quantitative Structure-Activity Relationship (QSAR) are important theoretical tools for ADR prediction. Pharmacophores refer to the spatial arrangement forms of structural features in drug molecules or bioactive molecules that play an important role in activity; in traditional structural modeling methods, pharmacophore models can simulate the active conformation of ligand molecules through conformational search and molecular superposition, based on which the possible interaction modes between receptors and ligand molecules can be inferred and explained. Although SAR, QSAR, and pharmacophore models have revealed the association between drug structure and biological activity to a certain extent, they typically rely on rules and prior knowledge, with prediction of ADR types limited to predefined sets, often rendering it difficult to predict unknown ADRs. To break through these limitations, modern computational methods have shifted towards using machine learning and fusing multi-dimensional feature information of drugs to achieve significant improvements in prediction performance, such as chemical features (molecular structure), biological features (interactions of targets, pathways, cellular components [1, 2]), and phenotypic features (observable indicators such as therapeutic effects and adverse reactions[3]). However, this strategy relying on multi-dimensional features is impeded by a severe "cold-start" problem when predicting potential adverse reactions for novel drugs. The "cold-start" problem stems from the incompleteness and scarcity of data in the new drug R&D stage; for a candidate drug that has just been designed or synthesized, the only determinable information is its molecular structure, whereas its corresponding target information, pharmacokinetic (PK/PD) data, cell line response, and even clinical phenotype data are completely missing or extremely sparse at this stage. Therefore, how to build an ADR prediction model that does not rely on scarce experimental data solely based on the molecular structure of drugs is the key to solving the ADR cold-start prediction. Graph Motifs, as key substructures representing the local chemical environment and topological characteristics of molecules, are considered the structural basis for inducing specific biological effects (including adverse reactions), providing a new research pathway for deeply mining ADR structure-activity relationships.

Furthermore, the ADR prediction task exhibits significant multi-label characteristics, meaning a single drug is often associated with multiple adverse reaction categories, and known ADR types amount to as many as 2.59 million, a characteristic that poses substantial challenges to existing modeling methods. On one hand, semantic co-occurrences or latent dependencies often exist between labels, and simple independent modeling fails to effectively capture this relationship; on the other hand, the universal label space is vast and sparse, exacerbating the learning difficulty and generalization challenge of the model; more importantly, scenarios exist in real-world applications where some ADRs have not been discovered, posing new demands for the "open-ended prediction" capability of models, that is, to not only perform multi-label classification within the known label space but also to perform inference and prediction on unknown ADRs within the unknown label space, thereby transcending the limitations of traditional closed-world methods. Therefore, constructing a prediction framework capable of capturing complex relationships between labels in a universal space and adapting to an open label space holds substantial theoretical significance and practical value for ADR prediction modeling.

To address the aforementioned challenges, here we propose Graph-Motif feature fusion and Multi-Label Generation (GM-MLG), an open-ended prediction paradigm for ADRs, with main contributions summarized as follows:

1. Proposing a drug molecular representation learning method based on a dual-graph architecture. By obtaining fine-grained retrosynthetic motifs through dynamic fragmentation, constructing a dual-graph representation architecture and interactive fusion learning strategy spanning atomic level, molecular local level (i.e., motif), and molecular global level effectively solves the cold-start problem caused by the scarcity of multi-dimensional features and high-quality labeled data for new drugs, provides sufficient, open, and generalizable feature information support for the exploration and connectivity of reasoning pathways in ADR prediction, and simultaneously provides an interpretable and retrosynthetic new pathway for understanding ADR structure-activity relationships and drug molecular structure optimization.

2. Proposing for the first time an innovative prediction scheme that transforms ADR prediction from traditional Multi-Label Classification (MLC) into multi-label generation based on a Transformer Decoder. Treating the ADR label set as a discrete token sequence, with a basic Transformer Decoder as the prediction model and molecular structural features as input, the prediction token sequence of ADRs is generated step-by-step in an autoregressive manner. This generative prediction scheme fundamentally resolves the curse of dimensionality faced by traditional MLC methods under the universal ADR label space, averting task failure caused by high model complexity and sharp decline in generalization performance; moreover, under the premise of fixing the label set to ensure limited resource overhead and reduce label space sparsity, it dynamically expands the prediction space to tens of thousands of ADRs, transcending the closed nature of MLC drug prediction methods and innovatively realizing large-scale, open-ended prediction capabilities for potential unknown ADRs; furthermore, by adopting positional embedding encoding and the Transformer Decoder multi-head attention mechanism, it explicitly models and captures in parallel the complex long-range dependencies and co-occurrence relationships among multiple labels in the large-scale ADR label space, significantly improving ADR prediction performance.

3. Analyzing the contribution degree of different motifs to ADR prediction, revealing the non-simple linear additive structure-activity relationship existing between molecular motifs with retrosynthetic attributes and ADRs, thereby, on the basis of the above two contributions providing innovative methodological support for the "prospective systematic research discovery" of toxic ADRs of drugs (especially new drugs), this contribution further provides scientific basis and strategic guidelines for "molecular structure optimization design and toxic ADR regulation" of drugs.

2. Related Work

## 2.1. Multi-label Classification Learning

Multi-Label Classification (MLC) is a supervised learning problem where each instance may be associated with multiple labels. ADR prediction is a typical MLC problem because a single drug often elicits multiple side effects. Moreover, compared with general multi-label datasets, the label scale in ADR tasks is vastly larger, with known ADR types amounting to as many as 2.59 million, belonging to a ten-thousand-dimensional large-scale multi-label space; for example, the commonly used dataset SIDER (Side Effect Resource) records 5,868 types of ADRs, and DrugBank records tens of thousands of ADR-related information. Furthermore, the ADR label space distribution universally exhibits characteristics of sparsity and long-tail distribution.

Traditional MLC models often adopt the Binary Relevance (BR) strategy, which decomposes the multi-label learning task into multiple independent binary classifiers to predict whether a specific drug elicits a specific side effect. However, this method completely overlooks the correlations between multiple labels; for example, the adverse reactions vomiting and diarrhea exist with a certain correlation, but the BR strategy models the two separately, failing to capture and utilize the association and dependency relationship between them, leading to severely constrained prediction performance.

Classifier Chains (CC) introduce the concept of association dependency modeling, transforming the multi-label problem into chain-like binary classification tasks to model the association and dependency relationships among labels by using the output of one classifier as the feature for the next classifier. However, the CC method suffers from two problems. First, its computational complexity scales polynomially with the increase in the number of labels; for example, experiments on the Corel5k dataset (374 labels) demonstrate that the training time increased by approximately 50% compared to BR [4]. Second, CC is relatively sensitive to the label permutation order; if a fixed order is adopted, early classification errors may propagate layer-by-layer in the chain structure, severely affecting prediction performance.

Ensemble of Classifier Chains (ECC), addressing the adverse effects brought by multi-label order, proposes constructing multiple random label chains and integrating their prediction results to improve stability and generalization capability. However, the increase in computational overhead of ECC is more significant, and since each chain only focuses on local label dependency relationships, the modeling capability for complex label dependency structures on the whole remains limited. More importantly, in ADR large-scale label scenarios, ECC may even fail to complete training due to memory or time limitations. Taking the Delicious dataset as an example (983 labels, 16,105 samples), ECC failed to run successfully due to excessive resource consumption [5].

To address the deficiencies of ECC in label dependency modeling, researchers have begun to explore constructing novel MLC frameworks in recent years. For instance, Wang et al. [6] adopted a cross-fusion method, integrating image representations using Convolutional Neural Network (CNN) and label co-occurrence embeddings using Graph Neural Network (GNN) for multi-label ADR prediction; Wang et al. [7] improved the prediction task by using Swin Transformer and two-layer Graph Convolutional Network (GCN) to generate image representations and obtain label co-occurrence embeddings, respectively; Luo J et al.[8] proposed pADR, a drug adverse reaction prediction method integrating multi-source data to perform personalized prediction of ADRs, where their method used a variety of data features and could appropriately fuse them. Distinct from the aforementioned construction of novel MLC frameworks, Qian et al. [9] introduced a Long Short-Term Memory (LSTM) model incorporating attention mechanisms, directly feeding all known

ADRs of each instance as text sequences into the model for training and learning; by capturing dependencies between texts at different distances within the sequence to model ADR association relationships, it thereby predicts output unknown ADRs; although this study only provided experimental results on one dataset with a prediction accuracy of 58.08%, it offered a novel attempt for the modeling challenge of ADR multi-label dependencies.

Furthermore, the MLC methods for ADR are inherently "closed-world" prediction frameworks. Specifically, assuming the scale of real ADRs is denoted as $N$ types (currently known ADRs reach as high as 2.59 million), constrained by model complexity, the label set size of MLC is generally limited to no exceed $n$ (usually 200 label categories, corresponding to 200 types of ADRs), and a substantial gap exists between the two; for a specific drug, it is entirely possible to exhibit some ADRs that have appeared in other drugs (i.e., exist in $N$) but have not yet been discovered and recorded in the $n$ label set; however, MLC methods cannot transcend the limited and fixed "closed" ADR label set (i.e., $n$) to explore the open potential ADRs of the drug (located in the $N - n$ label set), that is, MLC lacks the open prediction capability for potential unknown side effects of drugs. Particularly for new drug R&D, new drugs often lack complete side effect records; sole reliance on closed-world prediction may overlook certain severe toxic side effects, resulting in huge medication safety risks and substantial losses in new drug R&D costs; therefore, "open-ended prediction" methods for ADR warrant urgent investigation.

## 2.2 Representation Learning Based on Molecular Structural Features

In ADR prediction, based on distinct data sources, representation learning methods can be broadly categorized into multi-dimensional feature methods and single-dimensional feature methods. Multi-dimensional methods integrate multi-source information such as drug targets, phenotypes, and pharmacokinetics; however, in the "cold-start" scenarios of new drug R&D, the scarcity of multi-dimensional feature data and high-quality labeled data often impedes prediction capabilities. Molecular structure serves as an explicit intrinsic feature known for every candidate drug; consequently, prediction methods based solely on the single dimension of molecular structure have garnered significant attention due to their universality and cost-effectiveness. Within this research trajectory, molecular representation studies predominantly revolve around global features, local features, and the integration of both, aiming to achieve high-performance representation learning under conditions of limited information.

### 2.2.1 Global Features of Molecular Structure
1. Structural Identifiers

To facilitate computational processing and modeling, it is usually necessary to transform and represent molecular structures into recognizable text or numerical forms. International Chemical Identifier (InChI) and Simplified Molecular Input Line Entry System (SMILES) are the two most commonly used molecular structural identifiers, also known as molecular linear notations. They encode molecular structural information through compact string formats, providing a foundation for subsequent feature extraction and descriptor calculation, and have found widespread application in the fields of cheminformatics and drug design.

SMILES is a representation specification that compresses molecular structure into a one-dimensional ASCII string sequence. Researchers utilize this serialized input to learn molecular features by processing text-based network architectures. Bender et al. [10] made the first attempt to predict hundreds of classes of side effects solely from drug chemical structure SMILES, and demonstrated the feasibility of using drug chemical structures to predict side effects. Subsequently, extensive studies[11-14] have been conducted based on SMILES to perform representation learning and enhance ADR prediction performance.

The byte length of InChI is usually longer than that of SMILES, which is attributed to the fact that it contains richer and more detailed chemical information, such as molecular structural layers, isotopic distribution, stereochemistry, and charge states. Although this long format is not as concise as SMILES, it can effectively eliminate ambiguity and ensure high-precision description of complex molecules. Furthermore, the hierarchical structure of InChI provides advantages for database applications, enabling molecular structures to be retrieved and compared more reliably and consistently. Therefore, InChI is applicable to scenarios demanding precise chemical descriptions.

2. Molecular Graphs

String-based representations such as SMILES and InChI are widely employed, but their linearization process fails to effectively preserve the complete topological and geometric information of molecules. Consequently, researchers have increasingly shifted towards graph structures that can more directly encode atom and bond connectivity [15, 16]. Molecular graphs employ nodes to represent atoms with specific chemical properties and edges to represent different types of chemical bonds; moreover, edges are capable of encoding physicochemical properties such as bond strength and length, thereby comprehensively reflecting molecular topology and spatial structure. Therefore, molecular graphs exhibit distinct advantages in capturing intra-molecular interactions and conformational features, preserving richer structural details, rendering them more suitable for molecular modeling and property prediction.

Given the natural alignment between molecular structure and graph data structure, researchers have initiated utilizing technical methods related to Graph Neural Networks (GNN) to perform representation learning of molecular features. GNN is a deep learning model tailored for processing graph-structured data, capable of performing operations directly on graphs and updating atom representations by aggregating neighbor node information, thereby effectively learning molecular topological features. Zhao et al. [17] proposed GSL-MPP, which extracts molecular representations via GNN, constructs molecular similarity graphs, and conducts graph structure learning to fuse internal and external molecular information, ultimately used for molecular property prediction. Zhao et al. [18] proposed a method using contrastive loss and cross-attention to align and fuse graph and text features in the embedding space. Lee et al. [19] proposed a hybrid deep learning model composed of a Graph Convolutional Neural Network (GCNN) with Inception modules and a Bidirectional Long Short-Term Memory (BiLSTM) recurrent neural network, where the former is used to learn drug molecular features more efficiently, and the latter is used to link drug structures with their associated ADRs. Bongini P et al. [20] proposed a novel model GNN–MGSEP based on molecular graphs describing drug structures to predict drug ADRs. Deac et al. [21] proposed a Graph Neural Network (GNN) based on a general attention mechanism, utilizing ADR types and drug molecular structures to predict potential ADEs arising from multi-drug combinations. Yadav M et al. [22] proposed a new model named GTransfNN (Graph-based Transformer Neural Network), which utilizes graphs with Transformers to analyze the molecular structure of drugs, aiming to predict 27 ADR categories based on system organ classes. In recent years, pre-training frameworks based on Graph Self-Supervised Learning (Graph SSL) [23, 24] have developed rapidly and performed well in molecular property analysis tasks. These ADR prediction methods generally focus on global graph structures, often neglecting the study of unique substructures, such as rings and functional groups.

2.2.2 Molecular Substructure Features

1. Molecular Fingerprints

Molecular Fingerprint is a widely employed molecular substructure representation method, such as Molecular Access System (MACCS fingerprints [25]), PubChem fingerprints [26], Extended Reduced Graph (ErG) fingerprints [27], and Extended Connectivity Fingerprints (ECFPs [28]). Molecular fingerprints aim to represent the presence or absence of various substructures in molecules, usually represented in the form of sparse vectors. Their construction methods are mainly categorized into two major classes: one class matches substructures in molecules with a predefined expert substructure set; the other class performs algorithmic enumeration and hash encoding on molecular substructures.

Duvenaud et al. [29] utilized molecular fingerprints as input to acquire molecular features and proposed a CNN operating directly on graphs for molecular property prediction. Sanjoy Dey et al. [30] used the ECFPs algorithm to extract molecular features and employed a deep learning framework to generate neural fingerprints based on ECFPs for ADR prediction. Li H et al. [31] constructed a deep drug representation module combining PubChem molecular fingerprints with biological features of drug molecules to predict ADRs.

Molecular fingerprints primarily focus on the substructure features of molecules while having limited consideration for the physicochemical properties or the full molecular three-dimensional conformation, leading to limitations in describing complex molecular structural features [32]. Furthermore, different types of fingerprints possess distinct advantages and limitations: MACCS fingerprints, PubChem fingerprints, and ErG fingerprints are mainly based on predefined substructures or functional group sets, therefore exhibiting low sensitivity to novel substructures not in these sets; whereas Morgan Fingerprint (Morgan FP), as a widely used variant of ECFP, encodes molecular structures through atomic neighborhood expansion, possessing the capability to describe unknown functional groups to a certain extent, but may suffer from information loss due to bit collisions during hash mapping operations.

2. Motifs

To compensate for the aforementioned limitations such as the absence of chemical semantics in molecular fingerprints, researchers have begun to explore substructures represented by Motifs, which possess greater chemical significance. Serving as key substructure units, motifs are important substructure patterns that appear frequently and possess semantic meaning, sometimes associated with the concept of pharmacophores [33]. Specific motifs not only reflect chemical structural features but also determine molecular biological activity. Motif analysis can elucidate the influence of key structural elements on efficacy and target binding, providing biological guidance for mitigating ADR.

To extract these meaningful motifs from molecules, researchers have designed various fragmentation strategies, mainly categorized into two classes: irregular fragmentation and rule-based fragmentation based on retrosynthetic analysis. In irregular fragmentation, simple methods such as retaining only the most basic bonds and rings, although simple and universal, lack semantic expression of functionality and reactivity; whereas eMolFrag [34] performs fragmentation through electron distribution and topological structure, which, although similarly independent of chemical reaction rules, can effectively reveal intrinsic chemical characteristics such as bond polarity and reactive sites in molecules. In contrast, rule-based fragmentation methods based on retrosynthetic analysis possess greater chemical interpretability, with their predefined cleavage rules corresponding to synthetic reactions in the real world; for example, RECAP [35] defines 11 types of cleavable bonds related to chemical reactions, which can be used to decompose complex molecules into chemically meaningful fragments; BRICS, on the other hand, is based on finer-grained rules, extending the 11 types of cleavable bonds to 16 types to obtain fine-grained motifs.

The dynamic application of these rules can extract common and potential novel motifs, enabling models to learn richer local structural features, thereby enhancing the generalization capability for molecules containing new structural units. Inae et al. [36] proposed a novel graph pre-training solution MoAMa based on a motif-aware attribute extraction strategy, utilizing motif structural knowledge to learn long-range inter-motif dependencies. Zhang et al. [37] proposed a Motif-oriented Representation Learning (MOTOR) method with topological refinement for DDI prediction, utilizing multi-grained motif information and topological structures of DDI networks to effectively capture internal motif structures, local motif contexts, and global motif semantics. Furthermore, Zhao et al. [38] proposed DEPOT, employing the chemical decomposition method RECAP to partition drug molecules into semantic motif trees and designing a structure-aware Graph Transformer to capture motif associations, thereby developing an innovative drug recommendation framework from a motif-aware perspective. To date, motif-based ADR prediction research has not been reported.

2.2.3 Representation Learning Methods Combining Global and Local Molecular Structural Features

Methods based on global features directly utilize the overall structural information of drug molecules for modeling, capable of capturing the global topological structure and long-range dependencies of molecules. In contrast, local feature methods prioritize local structural information such as key functional fragments and motifs within molecules, structures that are often closely associated with the occurrence of ADR. However, sole reliance on global features may overlook the influence of key local structures, whereas sole reliance on local features may lack overall molecular structural information. Therefore, increasing research tends to combine the two to capture both macroscopic and microscopic information.

To effectively integrate these two types of features, researchers have begun to explore hierarchical or multi-scale fusion methods. Zhu et al. [39] proposed the Hierarchical Information Graph Neural Network (HiGNN), which uses BRICS to segment molecular graphs into fragments, and then processes both the original molecular graph and its fragments using GNN to generate hierarchical representations of molecules for subsequent molecular property prediction. Zhang et al. [40] proposed Motif-based Graph Self-Supervised Learning (MGSSL), which utilizes the retrosynthesis-based algorithm BRICS and additional rules to design a universal generative pre-training framework based on motifs, implemented through breadth-first or depth-first methods, where GNN is required to perform topology and label prediction, and multi-level self-supervised pre-training is used to process multi-scale information in molecular graphs. Although these methods can respectively extract overall topological information or key functional motifs of molecules, in the feature integration process, a lack of effective interaction mechanisms in the fusion strategies for the two types of information will render the model unable to capture cross-scale structure-function associations within molecules, impeding the further improvement of model prediction performance.

3. An Open-Ended Framework for Adverse Drug Reaction Prediction via Graph-Motif Feature Fusion and Multi-Label Generation

This paper proposes GM-MLG, an open-ended paradigm for ADR prediction based on graph-motif feature fusion and multi-label generation: to address the cold-start challenge arising from the scarcity of multi-dimensional feature data and high-quality annotated data for drugs (especially novel drugs), GM-MLG constructs a dual-graph representation architecture covering the atomic level, the molecular-level local scope (based on fine-grained motifs obtained via dynamic segmentation using the BRICS algorithm and additional rules), and the molecular-level global scope, based on molecular structure, an intrinsic inherent feature explicitly known for every candidate drug; by learning information at each level through interactive fusion strategies, it provides sufficient, open, and generalizable molecular structural feature support for the exploration and connectivity of reasoning pathways in ADR prediction; targeting the modeling difficulty in capturing closed label sets and co-occurrence dependencies in MLC, GM-MLG proposes for the first time an innovative prediction scheme transforming ADR prediction from traditional multi-label classification into multi-label generation based on a Transformer Decoder, where under this generative paradigm, the model treats the ADR label set as a discrete token sequence, adopting molecular structural features as input, with positional embedding encoding explicitly modeling and capturing dependencies and co-occurrence relationships within the large-scale label space, generating ADR prediction token sequences step-by-step in an autoregressive manner, dynamically expanding the label prediction space, and constructing an open-ended prediction mechanism.

The overall framework of GM-MLG is illustrated in Figure 1: it generally comprises three types of nodes, atom nodes, molecule nodes, and motif nodes (important substructure patterns with biological significance that appear repeatedly in molecules, identified and extracted from molecular structures); the molecular graph provides an atomic-level feature encoding system; the molecule-motif association graph constructs an association interaction mechanism for local and global feature information at the molecular level, dynamically learning the contribution weights of different neighbor nodes to the central node representation via Graph Attention Networks (GAT), elucidating potential substructure commonalities existing between different molecules, including known motif commonalities and, more importantly, novel motif commonalities, exploring and connecting possible reasoning pathways for side effect prediction of new drug molecules; finally, the molecular graph and motif-molecule association graph complete dual-graph feature fusion via Multilayer Perceptron (MLP); the fused molecular structural features serve as the sole input for the Transformer Decoder, with positional embeddings encoding the ADR discrete token sequence, generating adverse drug reactions continuously in an autoregressive manner, completing open-ended prediction.

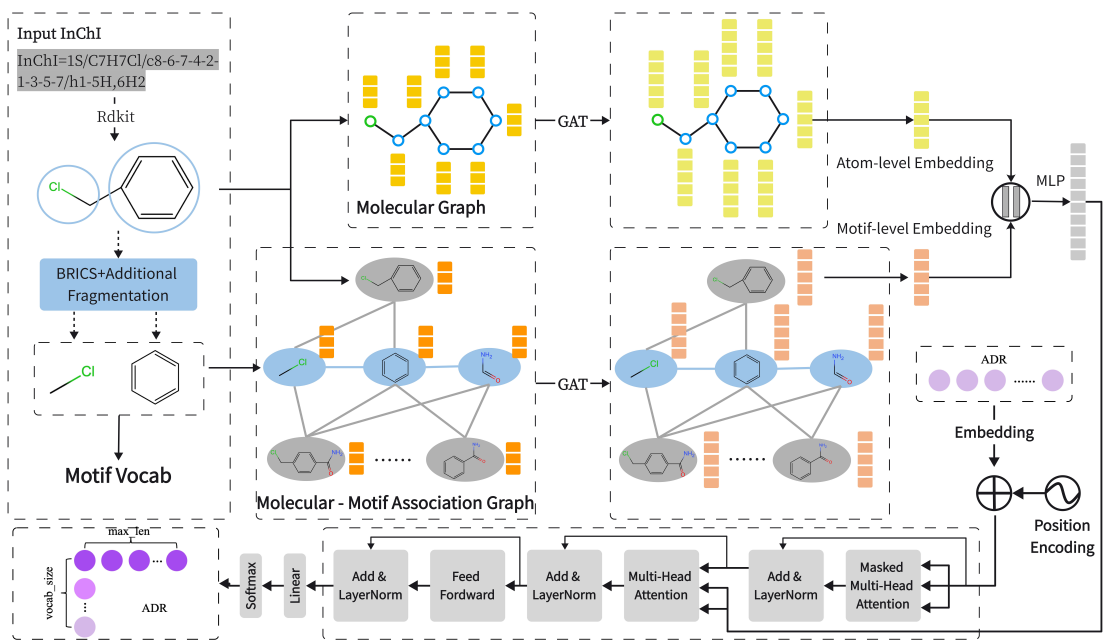

Figure 1 Overall Framework of GM-MLG

### 3.1. Molecular Graph Representation

In most public data sources, bioactive drug molecules are stored in the form of InChI or SMILES texts; to accurately and appropriately extract the structural features and physicochemical properties of drug molecules, it is necessary to transform bioactive drug molecules represented in InChI or SMILES forms into graph data structures processable by neural networks. The design rationale of InChI is to transform the structural features of compounds into a detailed and ordered text string, thereby enabling the precise description of their molecular composition and stereoconfiguration through this identifier regardless of the complexity of the compounds. As described in Section 2.2.1, InChI contains richer and more detailed chemical information relative to SMILES, to ensure high-precision description of complex molecules. The overall physicochemical properties of drug molecules are essentially the result of the interaction and integration of atoms representing individual physicochemical properties; therefore, this section utilizes the physicochemical characteristics of atoms as basic elements to construct an atomic-level feature encoding system for drug molecules based on InChI.

We represent each molecule as a molecular graph, where nodes and edges represent atoms and chemical bonds, respectively. Node features are 9-dimensional, including atomic number, degree of connectivity, formal charge, chirality, number of hydrogen atoms, hybridization state, aromaticity, atomic mass, and number of radical electrons. Edge features are 3-dimensional, including bond type, bond stereochemical property, and conjugation property. On this basis, edge weight information is introduced, and GAT is adopted to learn atomic-level feature embeddings.

$$\vec{h}_j = \vec{x}_j \, W \tag{1}$$

$$\vec{e}_{ij} = LeakyReLU\left(\vec{a}^T[\vec{h}_i \parallel \vec{h}_j \parallel \vec{E}_{ij}]\right) \tag{2}$$

$$\alpha_{ij} = softmax_j(\vec{e}_{ij}) = \frac{exp(\vec{e}_{ij})}{\sum_{k \in \mathcal{N}_i} exp(\vec{e}_{ik})} \tag{3}$$

In Equation 1, the edge feature $\vec{E}_{ij}$ represents the connectivity relationship between node $i$ and node $j$, which can provide additional information to enhance the association between nodes. Before calculating the attention score, we integrate the edge feature $\vec{E}_{ij}$ with the node feature $\vec{h}_j$ to obtain a richer neighborhood representation, where $W$ is a shared weight matrix used to linearly transform the node features, mapping the input features to the expected output feature dimension of the multi-head attention; $\|$ denotes the vector concatenation operation. In Equation 2, $\vec{e}_{ij}$ represents the calculated attention score, and this formula reveals the relative importance of the neighbor node $j$ to the central node $i$. $\vec{a}$ is a learnable weight vector that assigns importance to different neighbors of $i$, enabling the model to highlight specific neighbor node features that are more relevant to the task. Equation 3 employs the $\text{softmax}_j(\cdot)$ activation function to normalize the attention scores calculated in the previous step to obtain the attention weights $\vec{e}_{ij}$, revealing the importance of neighbor nodes to the current node.

After obtaining the attention weights for each neighborhood, normalization is performed, followed by a weighted summation of the node features within these neighborhoods. The aforementioned process results in the output of the node attention module, which represents the final updated node feature $\vec{h}_i'$. The updated node feature representation is shown in Equation 4, where $\sigma(\cdot)$ denotes the non-linear activation function, and $U$ is the edge feature update matrix.

$$\vec{h}_i' = \sigma\left(\sum_{j \in \mathcal{N}_t} \alpha_{ij} \vec{h}_j U\right) \tag{4}$$

To stabilize the training process of the GAT and enhance the expressive power of the model, the GAT employs multiple independent attention mechanisms to calculate hidden states, where each attention mechanism is referred to as a head, and each attention head possesses its own trainable parameters. Finally, the outputs of each attention head are aggregated via weighted averaging to obtain the final node features, as shown in Equation 5, where $K$ represents the number of heads.

$$\vec{h}_i' = \sigma\left(\frac{1}{K}\sum_{k=1}^{K} \sum_{j \in \mathcal{N}_i} \alpha_{ij}^k \vec{h}_j U^k\right) \tag{5}$$

3.2. Construction of Molecule-Motif Association Graph

In the aspect of combining global and local features of molecules, existing research methods are capable of respectively extracting overall topological information or key functional motifs of each drug molecule; however, in the feature integration process, the absence of effective interaction mechanisms renders the fusion strategies of the two types of information difficult to capture internal cross-scale structure-function associations between different drug molecules, thereby limiting the improvement of model prediction performance. To address this issue, this study proposes constructing a Molecule-Motif Association Graph, which not only performs unified modeling of the overall structure and key motif information of a specific drug molecule, realizing the synergistic extraction of molecular structure global and local representations, but also constructs an association interaction mechanism between molecule nodes and motif nodes; by dynamically learning the contribution weights of different neighbor nodes to the central node representation via GAT, it reveals latent substructure commonalities existing among different drug molecules, including known motif commonalities and, more importantly, novel motif commonalities, suggesting possible reasoning pathways for ADR prediction.

### 3.2.1. Fine-grained Retrosynthetically Interesting Motif Vocabulary

In molecular graphs, motifs are defined as subgraphs that appear frequently across multiple classes of molecules and possess statistical significance (as detailed in Section 2.2.2). The motif vocabulary is a collection of valid subgraphs extracted from given molecular graphs, serving as the foundation for understanding and characterizing molecular structures. Constructing a motif vocabulary necessitates identifying and recording all significant substructures, systematizing structural analysis and representation, thereby facilitating an in-depth understanding of structural features and functional relationships.

Distinct from traditional molecular fingerprints that map molecular topological environments into fixed-length binary vectors via hashing, resulting in the loss of structural semantic information and irreversibility, to ensure that motifs possess chemical property intuitiveness and convey semantic information, this paper adopts a chemically inspired molecular fragmentation method, transforming the molecular graphs of all drugs in the training set into motif-based graphs. As described in Section 2.2.2, current fragmentation methods mainly comprise irregular and regular types; although irregular algorithms possess advantages in revealing physicochemical essences such as molecular bond polarity and reactive sites by virtue of their flexibility independent of predefined reaction lists, to ensure that the extracted substructures possess the retrosynthetic attributes and interpretability required for drug design, this paper adopts the BRICS algorithm (Breaking of Retrosynthetically Interesting Chemical Substructures [41]) based on retrosynthetic concepts. From a retrosynthetic perspective, drug molecules are not random accumulations of atoms, but are combinations connected by a series of standard chemical reactions (such as amide condensation, etherification), and the BRICS algorithm precisely identifies the chemical bonds (such as amide bonds, ester bonds) formed in these synthetic reactions, retrosynthetically reducing the molecules into structures with high synthesizability and replaceability. Compared with RECAP, BRICS extends the 11 types of cleavable bonds to 16 types, aiming to preserve molecular components with key retrosynthetic structures and functional content, such as aromatic rings. On this basis, this paper applies additional fragmentation rules to the motifs output by the BRICS algorithm [42]: (1) breaking the bonds between a ring and its substituent groups; (2) designating non-ring atoms with three or more adjacent atoms as new motifs and cleaving adjacent bonds. This process reduces graph complexity while retaining basic chemical information, ultimately obtaining a fine-grained motif vocabulary. As illustrated in Figure 2, the above process comprises two stages: Step 1, BRICS Fragmentation), applying the BRICS algorithm to identify and cleave specific chemical bonds in the molecule (indicated by red crosses), preliminarily decomposing the drug molecule into ring structures and chain scaffolds; Step 2, Additional Fragmentation, focusing on the larger non-ring linker substructures obtained after preliminary decomposition, applying additional fragmentation rules for secondary fragmentation to obtain finer-grained motifs. Ultimately, all substructure fragments with chemical semantics are constructed into the model's Motif Vocab.

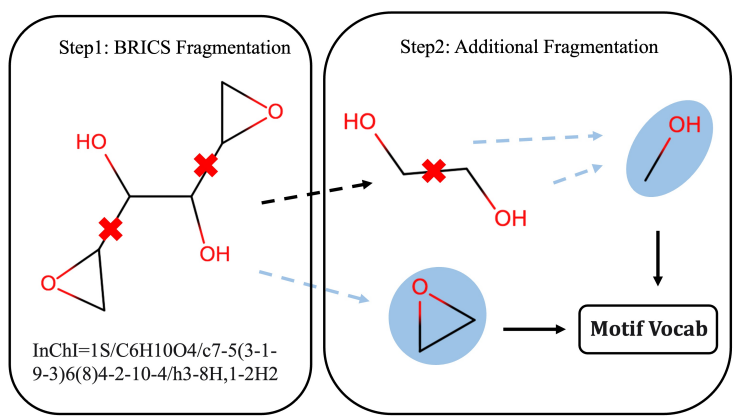

Figure 2 Decomposition Rules

To ensure a manageable vocabulary size, all duplicate motifs are removed. Certain motifs may appear in the majority of molecules; these ubiquitous motifs contribute limited informative value to molecular representation. To ensure the discriminability of molecular motifs and mitigate the interference of high-frequency non-informative features, the Term Frequency-Inverse Document Frequency (TF-IDF) algorithm is adopted. Specifically, Term Frequency (TF) measures the frequency of occurrence of a motif within a single molecule, whereas Inverse Document Frequency (IDF) represents the reciprocal of the number of molecules containing that motif. We take the average of the TF-IDF values of all molecules containing a specific motif as the TF-IDF value for that motif, so as to comprehensively reflect the overall importance of the specific motif, avoiding the stochastic influence of individual molecules, while simultaneously simplifying calculations and facilitating the comparison of importance among different motifs.

3.2.2. Construction of Motif-Molecule Association Graph

Based on the motif vocabulary, we constructed a molecule-motif association graph containing motif nodes and molecular nodes. Each motif node represents a motif in the vocabulary, and each molecular node corresponds to a molecule. Two types of edges are established between these nodes: a molecule-motif edge is added between a molecular node and the motif node it contains; if two motif nodes share at least one atom in any molecule, we add a motif-motif edge between the two motifs.

Distinct from methods relying on a single molecular graph, this modeling based on the molecule-motif graph is capable of enabling the sharing and transfer of substructure information across molecules: motif nodes serve as "bridges" in the graph connecting multiple molecules, facilitating information interaction between molecules through the message passing mechanism of GNN. From a biological perspective, motifs, functioning as pivotal determinants of molecular properties and functions, are often closely associated with drug activity and adverse reactions. Constructing the molecule-motif graph not only enables the capturing of intra-molecular structural features but also reveals potential commonalities among different molecules. More importantly, this molecule-motif graph structure offers significant advantages for "open-ended prediction"; the model leverages combinations of known motifs and emerging motif structures as hub nodes, establishing all association paths between established drug molecular structures and new drug molecular structures, providing potential plausible inference pathways for unknown side effects corresponding to new drug molecular structures.

The constructed Molecule-Motif Association Graph comprises all molecules in the training set

and all motifs in the vocabulary. Figure 3 illustrates an example of the local structure of the Molecule-Motif Association Graph; taking the Benzyl Chloride molecule node as an illustrative instance, dynamic fragmentation yields two motif nodes, which not only constitute the basic structural units of Benzyl Chloride but also appear in other molecules, thereby linking multiple molecule nodes to shared motif nodes. Furthermore, the graph also includes a motif node derived from other molecules; although this motif does not belong to Benzyl Chloride, it is associated with it via a motif-motif edge, demonstrating the role of motifs serving as cross-molecular bridges.

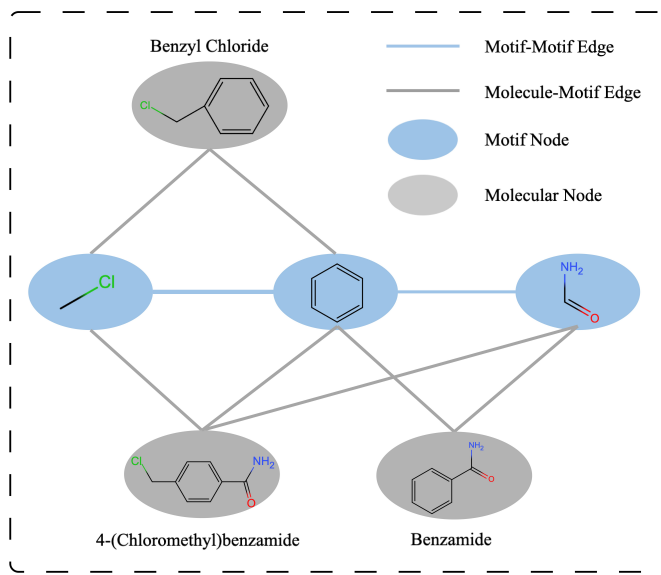

Figure 3 Molecule-Motif Association Graph

Since different motifs have different impacts, different weights are assigned to edges according to the terminal nodes of the edges. Specifically, for edges connecting motif nodes and molecular nodes, the TF-IDF value of that motif is used as the weight.

$$TF - IDF_{ij} = TF_{ij} \times \log\left(\frac{N}{1 + DF_i}\right) \quad (6)$$

Where $TF_{ij}$ is the term frequency, representing the number of occurrences of motif $i$ in molecule $j$, $N$ is the total number of molecules, and $DF_i$ is the number of molecules containing motif $i$.

For edges connecting two motif nodes, the Pointwise Mutual Information (PMI) of co-occurrence information is used as the weight, which is a correlation measure widely used in information theory and statistics. Herein, edges with negative PMI are set to zero weight.

$$PMI(i,j) = \log \frac{p(i,j)}{p(i)p(j)} \quad (7)$$

Where $p(i,j)$ is the probability that a molecule contains both motif $i$ and motif $j$, $p(i)$ is the probability that a molecule contains both motif $i$ is the probability that a molecule contains motif $i$, and $p(j)$ is the probability that a molecule contains both motif $i$ is the probability that a molecule contains motif $j$.

### 3.3. Dual-Graph Feature Extraction and Fusion Module

This paper constructs a dual-graph feature extraction and fusion module spanning atomic level, motif level, and molecular global level, designed to learn interactive fusion representations of multi-scale feature information. This module processes two types of graph structures in parallel: original molecular graphs and the molecule-motif association graph.

On the one hand, the raw molecular graph takes atoms in the molecule as nodes and chemical bonds as edges, carrying atom topological structures, node features, and bond features, where this model extracts atom-level features through GAT, capturing structural information between atoms and combining bond information to enhance atom representation. On the other hand, we constructed a molecule-motif association graph containing all motif nodes and molecular nodes, where each node in the graph undergoes feature initialization: for motif nodes, one-hot encoding is used to construct feature vectors, where each motif node $i$ has a feature vector $X_i$ of length $|V|$, with $V$ representing the motif vocabulary, and based on the unique index of the motif in the vocabulary, the $i$-th dimension is set to 1 while the rest are 0; for molecular nodes, the bag-of-words model is adopted to construct feature vectors, treating each motif as a "word" and each molecule as a "document" to construct molecular node feature vectors. Based on node features, motif-level and molecule-level global feature embeddings are learned through GAT, achieving cross-molecule substructure information sharing and feature interaction.

Different from traditional pooling methods, this model adopts an atom-position-based serialization scheme as the input for the Transformer decoder. Specifically, for a molecule containing $V$ atoms, an input sequence of length $N$ is constructed. Each position consists of two parts of features: one is the atom-level feature extracted by the atom graph GAT, and the other is the motif-level and molecular global feature extracted by the molecule-motif graph GAT, wherein the global feature is repeatedly mapped at each atom position, thereby forming a position-dependent enhanced feature representation. At the same time, the model uses a dynamic mask based on the actual number of atoms to process variable-length sequences, which not only retains atom-level structural details but also incorporates motif-level global context information. Finally, the above position-dependent enhanced feature representations are input into an MLP, and then input into the multi-label generation model for ADR sequence generation.

3.4. Multi-label Generation Model

LSTM are an improved type of Recurrent Neural Network (RNN), which regulates the recursive transmission of hidden states through gating mechanisms, allowing sequence dependency information to accumulate in the temporal dimension, effectively alleviating the gradient vanishing problem of traditional RNNs and enhancing the model's ability to learn dependency relationships. However, since the calculation of LSTM still relies on the sequential transmission of information between time steps, long-range dependencies are prone to decay during multiple iterations, and there is still a risk of gradient vanishing or explosion when processing long sequences, which limits its ability to model long-distance dependencies. In contrast, the Transformer architecture effectively models the relative positional relationships of sequences by combining masked self-attention mechanisms with positional encoding. Positional encoding allows the model to directly access information at any position in the sequence during calculation without stepwise transmission, thereby more efficiently capturing long-distance dependencies. In autoregressive training, when calculating the representation of position $i$, the model introduces a mask to ensure that only information from position $j$ satisfying $j \leq i$ is considered, meaning the current position will not be affected by subsequent positions. This precise capturing ability for long-distance dependencies makes Transformer excel in processing complex token sequence tasks. Furthermore, distinct from methods such as RNN and LSTM, Transformer possesses the significant advantage of parallel computing, handling large-scale datasets more efficiently.

This study transforms ADR prediction from traditional MLC into an innovative prediction scheme based on Transformer Decoder multi-label generation. Specifically, the ADR label set is regarded as a discrete token sequence, and the known ADR set of each drug is represented as a sequence $X = r_1, r_2, r_3, \ldots, r_n$, where $r_i$ corresponds to an ADR label. This sequence serves merely as a serialized expression of an unordered label set, where its internal arrangement does not bear actual semantic order assumptions, aiming to utilize the autoregressive decoding mechanism to characterize the statistical co-occurrence and dependency relationships between ADR labels, rather than reflecting the timing or causal relationships of ADR occurrence. During the ADR sequence construction process, if the number of ADR labels exceeds the preset maximum length (max_len), truncation is performed; if insufficient, it is supplemented using padding tokens. It is worth noting that, distinct from natural language generation tasks, the multi-label generation output results of this study do not exhibit hallucination issues at the semantic level; the generation space of the model lies within the range of the pre-constructed ADR label vocabulary, where all prediction results are encoded label indices, and the generation process solely reflects the learning and modeling of label co-occurrence and dependency patterns.

The Transformer decoder is composed of stacked multiple decoding layers [43], where each decoding layer contains three sub-modules: masked multi-head self-attention mechanism, cross-attention mechanism, and feed-forward neural network. Among them, the masked self-attention module is used to model the context dependency of the ADR label sequence; the cross-attention module is used to introduce the multi-scale structural features of drug molecules. In the cross-attention mechanism, the drug molecule fusion features extracted by the GAT in Section 3.3 serve as the Key ($K$) and Value ($V$) of the attention mechanism, while the adverse reaction representation of the current decoder state serves as the Query ($Q$), enabling the model to dynamically aggregate structural semantic information when generating each ADR. Through this interaction mechanism, the model is able to efficiently learn the potential associations between drug structures and adverse reactions. The calculation of a single attention head follows the standard form as follows:

$$Attention(Q, K, V) = softmax\left(\frac{QK^T}{\sqrt{d_k}}\right)V \tag{8}$$

Where $Q$, $K$) and Value ($V$, and $V$ represent the query, key, and value vectors respectively, and $d_k$ is the dimension of the key vector. This mechanism calculates attention weights through dot products and performs weighted aggregation on value vectors, achieving selective fusion of information.

The Multi-Head Attention mechanism further enhances the model's ability to extract information from multiple semantic subspaces, and its calculation is as follows:

$$MultiHead(Q, K, V) = Concat(head_1, \ldots, head_h)W^O \tag{9}$$

$$head_i = Attention(QW_i^Q, KW_i^K, VW_i^V) \tag{10}$$

Where $h$ represents the number of attention heads, $W_i^Q$, $W_i^K$, $W_i^V$ are the linear transformation matrices corresponding to each head, and $W^O$ is the output linear transformation matrix.

The multi-head attention mechanism allows the model to learn different attention weights from multiple sub-representation spaces, thereby more comprehensively modeling the complex relationships within sequences as well as with drug structures. The output of each layer is further processed through residual connections, layer normalization, and feed-forward networks, and finally outputs the probability distribution of adverse reaction labels through a linear layer and softmax. In the training phase, the model adopts a teacher forcing strategy, which uses the ground truth label of the previous moment as the input of the current step to accelerate model convergence while preventing error accumulation, and improves training efficiency and performance; in the inference phase, the model relies entirely on the generated historical labels for stepwise autoregressive generation until a termination token is output or the maximum label length is reached.

4. Experiments
4.1 Experimental Overview
4.1.1 Construction of Experimental Dataset IADRSeq

MetaADEDB [44] is integrated from multi-source authoritative databases, whose data sources include three categories of structured drug side effect databases (CTD, SIDER, and OFFSIDES), as well as two major spontaneous adverse reaction reporting systems (US FDA Adverse Event Reporting System (US FAERS) and Canada Vigilance Adverse Reaction Online Database). On this basis, MetaADEDB aggregates structured knowledge, real-world reported information, and large-scale drug-ADR associations, and through unified drug identification, deduplication, and normalization processing, constructs a high-coverage, high-quality drug adverse reaction knowledge base, which is subsequently upgraded to MetaADEDB 2.0.

Currently, there are no public resources directly associating standardized InChI codes with ADR sequences, so this study autonomously constructed a dataset and named it IADRSeq (InChI-ADR-Seq). It is worth emphasizing that this dataset is the first benchmark dataset to establish an end-to-end sequence mapping between complete InChI data formats and ADR labels. The specific construction method is as follows: the dataset is sourced from public authoritative database resources MetaADEDB 2.0 and PubChem [45], taking the MetaADEDB 2.0 database as the core resource and PubChem as the supplementary data source, corresponding InChI detailed information retrieval and matching were performed according to the InChIKey recorded in MetaADEDB 2.0; after integration, the InChI codes of 8,481 drugs and their corresponding 13,191 drug adverse reaction labels were finally obtained, serving as the foundational dataset for GM-MLG experimental research; to adapt to the generative prediction paradigm of the GM-MLG architecture, this study further concatenated all ADR labels associated with each drug into a comma-separated text sequence, wherein each ADR label is regarded as a "word" (token) element in this sequence, and since all tokens originate from a pre-encoded label set, this generative representation does not involve hallucination problems at the semantic level. This strategy establishes a correspondence between each drug InChI structure and all its related ADR data. The finally constructed IADRSeq dataset contains 8,481 records, where each record consists of a drug InChI code and the corresponding ADR sequence.

The experiment was executed with independent repeated verification through 5 different random seeds. The splitting of the dataset was strictly performed centering on drugs: firstly, unique numbering and random shuffling were performed only on drugs in the dataset, dividing them into three mutually exclusive subsets including the training set, validation set, and test set according to a ratio of 8:1:1. The construction of each subset not only contained the partitioned drug nodes, but also synchronously retained the motif nodes connected to these drugs and their topological connections, thereby forming three independent training, validation, and test subgraphs, ensuring that there is no possibility of any data leakage in the dataset. The final experimental results are the average of the test set performance metrics in 5 independent experiments.

### 4.1.1 Experimental Environment and Performance Metrics

In the training of deep learning models, hyperparameters are an important part that cannot be ignored, as their different choices may produce significant impacts on model performance. The experiments in this paper were run on a computing platform equipped with an NVIDIA GeForce RTX 4060 8GB graphics card, testing the model separately using different hyperparameter settings, and finally obtaining the optimal hyperparameter combination, as shown in Table 1。

Table 1　Main Hyperparameter List

| Name | Value | Description |
| --- | --- | --- |
| gat_heads | 2 | Number of heads in multi-head GAT |
| epochs | 50 | Number of training epochs |
| batch_size | 64 | Batch size |
| num_layers | 3 | Number of Transformer DecoderLayer |
| d_model | 128 | Word vector dimension |
| max_len | 200 | Fixed number of ADR |
| vocab_size | 13191 | Total number of labels |
| lr_max | 1e-3 | Maximum value of self-adjusting learning rate |
| lr_min | 1e-5 | Minimum value of self-adjusting learning rate |

Figure 4 shows the interval distribution of drug ADR counts in the original IADRSeq dataset, wherein the bar chart displays the frequency distribution with intervals of 100, while the line chart represents the cumulative percentage of the intervals. Data shows that the number of ADRs presents significant long-tail distribution characteristics, and in order to avoid unnecessary computational overhead caused by processing ultra-long sequences, while ensuring the integrity of the vast majority of sample information, we set the maximum sequence length to 200 (max_len=200), and based on this threshold, 87.1% of drug samples (i.e., samples with ADR counts in the 0-199 range) are able to completely retain all their adverse reaction labels, where the IADRSeq-200 dataset constructed based on the above truncation strategy serves as the input for subsequent model training and evaluation, effectively avoiding sparsity and computational redundancy problems caused by long-tail data.

In the evaluation phase, regarding the sequence results generated by the model, this paper adopts a set-based post-processing and evaluation strategy. Firstly, the predicted sequences and ground truth sequences are cleaned, truncating sequences with EOS as the terminator, and removing padding as well as other special tokens (UNK), retaining only valid ADR labels. Given that the ADR label set possesses permutation invariance[46], meaning that different sequence arrangements can correspond to the same semantic set, this paper does not adopt bitwise alignment sequence matching methods, does not take distance proximity as the evaluation metric, but instead uses an order-independent set-level matching strategy: as long as a predicted label exists in the ground truth label set, it is determined as a true positive. Referring to the performance metrics of Precision, Recall, and F1 score by Yang et al. [47]in the SGM model, the calculations of all three are based on global true positive, false positive, and false negative statistics between the predicted set and the ground truth set, rather than sequence position alignment, thereby objectively evaluating the predictive performance of the model in the unordered multi-label space.

The calculation formulas for performance metrics are shown as follows, where TP, FP, and FN represent True Positive, False Positive, and False Negative respectively:

$$Precision = \frac{TP}{TP + FP} \tag{11}$$

$$Recall = \frac{TP}{TP + FN} \tag{12}$$

$$F1 = \frac{2 \times Precision \times Recall}{Precision + Recall} \tag{13}$$

On the basis of the former, this study will further conduct open-ended prediction performance and generalization performance evaluation, so as to comprehensively analyze the robustness of GM-MLG in open label spaces and new drug scenarios.

To evaluate the training feasibility and operational efficiency of the model under limited computational resources, this paper further gathered statistics on the experimental overhead situation. When running on a single NVIDIA GeForce RTX 4060 (8GB VRAM) GPU, the maximum video memory occupation of the GM-MLG model is about 7.8 GB, the time consumption for a single round of training is about 5 minutes, and the total training time is about 5 hours.

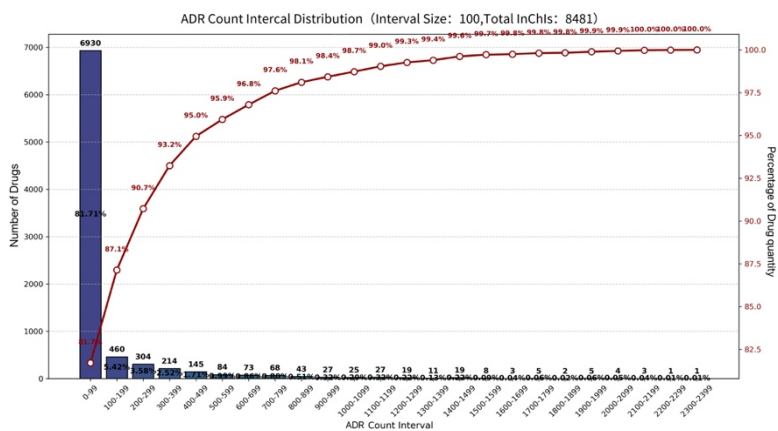

Figure 4 ADR Frequency Distribution Histogram

4.2 Baseline Comparative Experimental Study

To intuitively evaluate the performance advantages and limitations of the GM-MLG deep learning model, this paper conducts a detailed comparison with methods adopted in SOTA related studies.

a)  Lee's Model[19]:ADRs and drugs are connected using PubChem IDs associated with SIDER and DrugBank. The model consists of a Graph Convolutional Neural Network (GCNN) with an Inception module and a BiLSTM recurrent neural network, where a total of 149 SMILES strings are converted into graph drug molecular structures, and drug names along with their related graph drug molecular structures are input into the GCNN; while 23,101 drug ADRs along with drug names are treated as input text sequences and fed into the BiLSTM. The GCNN learns drug molecular features, the BiLSTM associates drug structures with their related ADRs, and the outputs of the two networks are concatenated, using a fully connected network to predict drug ADRs. Among them, the maximum sequence length is 208 characters.

b) IGN[] [48]: Evaluated on the SMM4H-2019 (2,160 sentences) and CADEC (7,597 sentences) datasets, this approach proposes an Interactive Graph Network (IGN) architecture that treats noun phrases generated via sentence parsing as candidate phrases to expand dictionary coverage. By constructing three types of word–phrase interaction graphs—namely, the Boundary graph (B-graph), Context graph (C-graph), and Semantic graph (S-graph)—and utilizing GAT for encoding, the model explicitly captures boundary constraints and contextual dependencies between words and candidate phrases. Finally, the word representations output by a BiLSTM are fused with the graph representations from the GAT and input into a Conditional Random Field (CRF) layer, which predicts and outputs the entity boundaries and labels of ADR via the BIO tagging scheme.

c) KGDNN[49]:The SIDER database serves as one of the data sources, containing 5,868 types of ADRs and 1,430 drugs, totaling 139,756 drug-ADR pairs. This model proposes a binary classification method utilizing knowledge graphs to predict ADR. This method constructs a knowledge graph (KG) containing six categories of entities, including drug IDs, ADR, target proteins, indications, pathways, and genes. Using Node2Vec and CBOW algorithms, graph nodes are embedded into a feature space, and then the embeddings of drugs and ADRs are concatenated and input into the KGDNN model to achieve ADR prediction.

d) HMGNN[50]+our's GLM:This experiment adopts the feature components and feature extraction model of HMGNN, as well as the sequence generation method proposed in this paper (i.e., GLM). HMGNN employs the most fundamental bonds and rings to construct the motif vocabulary, implementing the construction of the molecule-motif graph, and uses two GIN models to respectively implement molecular feature prediction. In this comparative experiment, building upon HMGNN and utilizing this dataset, the sequence generation method (GLM) is integrated to implement ADR prediction, to conduct a comparison with the proposed model.

e) GTransfNN[22]:Drug molecular structures are obtained from PubChem, and related ADRs are obtained from the SIDER dataset, classifying 1,427 types of ADRs into 27 System Organ Classes (SOC) according to the MedDRA classification, where this model is specifically developed for multi-label classification of drugs, utilizing graphs with Transformers to analyze the 2D structures of drug molecules, enabling it to predict 27 categories of ADRs based on the SOC framework.

f) ToxBERT[14]:Uses the SMILES strings of drugs as model input, learning through a Transformer encoder based on the ELECTRA architecture. In the pre-training phase, this model adopts Masked Token Detection (MTD) and Replaced Token Detection (RTD) tasks, enabling the discriminator to learn the distribution and structural patterns of drug molecules, thereby overcoming the dependence of traditional models on manually selected molecular fingerprints. In the downstream ADR binary classification prediction task, the sequence representation output by the model predicts through a fully connected layer, and utilizes Youden's Index to determine the optimal classification threshold.

g) ADENER[51]:This model is designed for Adverse Drug Event (ADE) entity extraction in social media texts, aiming to address the challenges of high noise, entity overlap, and discontinuity. It adopts a syntax-augmented grid-tagging architecture, modeling the ADE extraction task as a multi-label word pair classification, fusing multi-dimensional text features and capturing long-distance word pair dependencies through convolution capture layers, and integrating path-level dependency information through syntactic affine layers, so as to improve extraction accuracy in unstructured texts.

As shown in Table 2, the table lists the key performance metrics of the GM-MLG model and several representative previous research methods, including Precision, Recall, F1 Score, etc., used to comprehensively weigh and evaluate the overall performance of each method in the drug adverse reaction prediction task.

Table 2 Comparative Experiment

| Method | Precision | Recall | F1 | ADR Label Types |
|---|---|---|---|---|
| Lee's Model (2021) | 0.925 | 0.518 | 0.664 | 208 |
| IGN(2021) | 0.724 | 0.689 | 0.706 | - |
| KGDNN (2022) | 0.821±0.024 | 0.857±0.022 | 0.837±0.004 | - |
| HMGNN+ our's GLM (2022) | 0.9386±0.021 | 0.8578±0.044 | 0.8882±0.036 | 13191 |
| GTransfNN (2024) | 0.84 | 0.87 | 0.83 | 27 |
| ToxBERT (2025) | 0.664±0.034 | 0.913±0.064 | 0.766±0.014 | - |
| ADENER (2025) | 74.20 | 75.08 | 74.64 | - |

The data in Table 2 demonstrates that, compared to existing state-of-the-art models, the GM-MLG generative deep learning model proposed in this paper achieves significant improvements in aspects such as precision, recall, F1-score, and overall performance efficiency: the high precision and F1-score of the GM-MLG model fully indicate its high reliability when predicting ADR; the high value of recall implies that the GM-MLG model is capable of identifying more true adverse reaction cases, which is crucial for drug safety assessment; as the harmonic mean of precision and recall, the high value of the F1-score further proves the superiority of the GM-MLG model in balancing precision and recall.

KGDNN relies on knowledge graphs to integrate multi-source information for prediction, and its primary defect lies in the high dependency on prior knowledge; GTransfNN possesses only 27 coarse-grained ADR categories, unable to satisfy the demand for fine-grained labels in real-world ADR prediction; Although IGN effectively resolves the challenges of identifying non-standard entities and out-of-vocabulary words in social media text by constructing multi-dimensional interaction graphs, it fundamentally functions as a post-marketing surveillance tool that relies heavily on user-generated clinical feedback data; ADENER extracts information from social media texts and performs remarkably in scenarios with rich text corpora, but often faces data absence encountering cold-start challenge issues in the early stages of new drug R&D; Lee's Model combines GCNN and BiLSTM for multi-label classification, but its BiLSTM is only used to encode drug names and associated side effect sequences, having limited expressive capability in handling long-sequence dependencies; ToxBERT constructs features based solely on SMILES sequences, but one-dimensional sequence expressions cannot effectively preserve the complete topological and geometric information of molecules.

In the comparative analysis, KGDNN, GTransfNN, Lee's Model, and ToxBERT all belong to the traditional Multi-Label Classification (MLC) paradigm, essentially performing discriminative prediction within a predefined closed label space. Whether binary classification or multi-label classification methods, both address discriminative problems within a fixed label space; the closed discriminative paradigm has confined the model to predicting between fixed and limited label sets, i.e., ADR labels equal ADR categories; facing realistic open-world scenarios with tens of thousands of ADRs and constantly emerging novel side effects, traditional classification models fall short. Targeting this bottleneck, GM-MLG innovatively transcends the classification framework, reconstructing the task as a multi-label generation task through the multi-label generation paradigm. This paradigm shift renders the model no longer constrained by predefined label sets, thereby equipping it with the capability to dynamically predict unknown ADRs in a ten-thousand-dimensional space. In specific implementation, this study comprehensively considers the long-tail distribution characteristics of ADR labels and the time and memory overhead of the model, ultimately setting the maximum sequence generation length to 200 (as detailed in Section 4.1.1), this length not only covers the ADR quantity of the vast majority of drugs but also ensures the feasibility of training and inference under actual computational resources.

In practical applications, GM-MLG exhibits outstanding adaptability advantages in cold-start scenarios. Distinct from ADENER relying on social media texts or KGDNN relying on knowledge graph entity features, these methods often fail to predict effectively due to the absence of external resources when facing new drugs in the early stages of R&D that lack clinical records, target data, or pharmacokinetic information. GM-MLG, in contrast, based on the intrinsic features of drugs, can complete end-to-end inference from molecular structure parsing and dual-graph construction to ADR sequence generation, eliminating the dependence on external knowledge. With this generative paradigm based on molecular intrinsic features, the model still generates reliable ADR sequence predictions under conditions of no prior information, enabling it to provide immediate and effective safety alerts for drug R&D.

4.3 Open-ended Prediction Capability Evaluation

To further evaluate the open-ended generation capability of the GM-MLG model, this study identified ADRs for various drugs that were either truncated or not recorded by existing labels in this dataset, wherein the proportions of these two types of open predictions were approximately 7% and 80%, respectively. This study utilized external literature and multi-source datasets to conduct cross-validation on all prediction results, and the validation results were divided into two categories: some predictions were supported by existing evidence; while for prediction results where no evidence was retrieved, they were marked as "None". Specifically, the predicted ADRs are divided into three cases: first, ADRs that exist in the IADRSeq dataset but were not associated during training due to exceeding the set maximum sequence length (max_len=200), meaning they do not belong to the IADRSeq-200 dataset; second, ADRs that were never associated with the specific drug in the training set of IADRSeq, belonging to the global ADR category, and it can be confirmed through third-party authoritative data sources that the drug possesses this ADR; third, new drugs that belong to the SID dataset but not the IADRSeq dataset, aiming to perform prediction verification (zero-shot evaluation) on the zero-shot prediction capability for new drugs.

Tables 3, 4, and 5 respectively present the verification results of the three scenarios, where "Recorded ADR Types" represents the total number of ADR types for that drug in IADRSeq. As shown in Table 3, the model successfully predicted labels such as Nausea lost due to sequence

truncation for Allopurinol, which is commonly used to prevent chemotherapy complications (Tumor Lysis Syndrome); as shown in Table 4, regarding Nelfinavir Mesylate commonly used in anti-tumor drug repurposing research, the model accurately predicted Stomatitis triggered by it; while in Table 5, regarding the late-stage cancer palliative care drug Methylnaltrexone, the model successfully issued a warning for the potential risk of Tachycardia. The above empirical results strongly corroborate the open-ended generation advantages of GM-MLG; the model breaks through the closed limitations of traditional classification tasks, possessing the capability to predict and capture real and clinically significant potential ADRs in the ten-thousand-dimensional label space.

Table 3 ADR prediction on IADRSeq-200

| Drug | Training Dataset | Recorded ADR Types | Predicted ADR | Evidence Source |
|---|---|---|---|---|
| Restasis | IADRSeq-200 | 424 | Nausea | IADRSeq |
|  |  |  | Feeling abnormal | IADRSeq |
| Allopurinol | IADRSeq-200 | 784 | Pain | ADReCS |
|  |  |  | Nausea | ADReCS |
|  |  |  | Fatigue | IADRSeq |
| Nicotinic Acid | IADRSeq-200 | 635 | Nausea | IADRSeq |
|  |  |  | Pain | IADRSeq |
|  |  |  | Feeling abnormal | IADRSeq |

Table 4 Prediction of novel drug-ADR associations

| Drug | Training Dataset | Recorded ADR Types | Predicted ADR | Evidence Source |
|---|---|---|---|---|
| Nelfinavir Mesylate | IADRSeq-200 | 15 | Feeling abnormal | ADReCS |
|  |  |  | Stomatitis | ADReCS |
| Lanthanum Carbonate | IADRSeq-200 | 11 | Nausea | ADReCS |
|  |  |  | Tachycardia | None |
|  |  |  | Constipation | ADReCS |
| L-Cysteinylglycine | IADRSeq-200 | 1 | Fatigue | None |
|  |  |  | Chest discomfort | None |
| Prucalopride | IADRSeq-200 | 42 | Tachycardia | ADReCS |
| 4-Thiouridine | IADRSeq-200 | 1 | Nausea | None |
|  |  |  | Constipation | None |

Table 5 ADR prediction for new drugs

| Drug | Training Dataset | Recorded ADR Types | Predicted ADR | Evidence Source |
|---|---|---|---|---|
| Fosinopril | SID | 264 | Neuropathy | IADRSeq |
|  |  |  | Diarrhoea | None |
| Tarenflurbil | SID | 257 | Insomnia | IADRSeq |
|  |  |  | Feeling abnormal | IADRSeq |
| methylnaltrexone | SID | 19 | Tachycardia | FAERS |

## 4.4 Generalization Performance Evaluation

To evaluate the generalization capability of the GM-MLG model under different data distributions, this study further conducted verification on two independent public datasets, SID and ADRE, both of which originate from open authoritative ADR databases. Specifically, the SID dataset is based on the benchmark collection constructed by Timilsina[52] et al. by integrating SIDER and DrugBank, and after entity alignment and standardization processing according to the MedDRA system, it contains a total of 1,004 drugs and 5,599 standardized ADR terms, constituting 131,778 verified positive association samples, accounting for only 2.34% of all potential drug-ADR pairs; the ADRE dataset originates from the ADRECS database, and based on the screening and reconstruction of the ADRECS database by Cheng [53] et al., original ADRs were summarized into 27 categories according to System Organ Class (SOC), finally forming 2,248 drugs, 27 categories of ADR labels, and 37,092 drug-ADR association records, accounting for approximately 61.11%.

The processing of the two datasets was maintained consistent with the processing of the IADRSeq dataset of the proposed model: treating drug ADR labels as sequences, truncating the first 200 non-zero labels, and padding the insufficient parts to the maximum length (max_len=200). The model maintained the same hyperparameter settings as the main experiment on both datasets, without performing any targeted fine-tuning, to verify its stability and adaptability across datasets.

Table 6 Generalization Performance Test Results

| Dataset | Precision | Recall | F1 |
| --- | --- | --- | --- |
| SID （1004 drugs） | 0.9359±0.0213 | 0.8553±0.0242 | 0.8909±0.0207 |
| ADRE （2245 drugs） | 0.9440±0.0235 | 0.8665±0.0219 | 0.8945±0.0291 |
| IADRSeq （8481 drugs） | 0.9575±0.0079 | 0.9048±0.0228 | 0.9251±0.0171 |

The generalization experimental results based on the two datasets, SID and ADRE, are shown in Table 6, where compared with the performance achieved on the model's dataset IADRSeq (8,481 drugs / 13,191 labels), the metrics of the model on the two new datasets experienced a slight decline, because the number of drugs in the SID and ADRE datasets is significantly fewer than that in IADRSeq, thereby limiting the model's ability to perform adequate feature fitting in the sparse drug-ADR association space to a certain extent, consequently leading to a slight decline in generalization performance. Despite being constrained by the scale of the datasets, the experimental performance still maintained a high level overall, especially exhibiting outstanding performance in terms of Precision. The results indicate that, when confronted with drug-ADR data of different scales and different label distributions, the GM-MLG model possesses good robustness and generalization capability.

## 4.5 Ablation Experimental Study
### 4.5.1 Feature Ablation

To systematically analyze the impact of each feature module and their combinations on model performance, this study designed multiple groups of ablation experiments.

For the molecular local feature module, this study selected four mainstream molecular fingerprints: MACCS fingerprint, PubChem fingerprint, Pharmacophore ErG fingerprint, and ECFP fingerprint, serving as independent feature vectors input into the MLP respectively. In addition, the three types of fingerprints, MACCS, PubChem, and Pharmacophore ErG, were concatenated along the 0-th dimension, and then uniformly encoded via a fully connected layer to construct a molecular fingerprint feature representation and input into the MLP, denoted as FP.

In terms of motif feature extraction, this section constructed a motif extraction control group that only retains the most basic rings and bonds, denoted as Motif_nobrics, so as to evaluate the impact of BRICS and additional rule cutting methods on extracted motif features.

Among model input features, PK information is considered to have potential value for ADR prediction. PK feature input items are constructed based on drug pharmacokinetic text descriptions from PubChem, DrugBank, and PKDB databases, covering 7 core parameters including Half-Life, Bioavailability, Absorption characteristics, Metabolism pathways, Protein Binding, Clearance, and Pathway information. These pharmacokinetic attributes reflect the Absorption, Distribution, Metabolism, and Excretion (ADME) processes of drugs in the body, which have significant meaning for drug efficacy assessment, safety research, and ADR prediction. This study added PK features to the ablation experiments to evaluate their impact on prediction performance.

Based on the above features, this study designed ablation experiments for multiple groups of feature input combinations, including single feature input and multi-dimensional feature fusion input, where the results of each experiment and evaluation metric values are detailed in Table 7 below.

Table7 Ablation Experiments

| Method | Precision | Recall | F1 |
| --- | --- | --- | --- |
| Mol | 0.9366±0.0244 | 0.8557±0.0541 | 0.8861 ± 0.0438 |
| MACCS | 0.9551±0.0138 | 0.8928±0.0351 | 0.9166±0.027 |
| ERG | 0.9548±0.0121 | 0.8922±0.0292 | 0.9166±0.0227 |
| PubChem | 0.9555±0.0081 | 0.8944±0.0240 | 0.9183±0.0177 |
| ECFP | 0.9487±0.0156 | 0.8775±0.0408 | 0.9051±0.315 |
| FP | 0.9497±0.0211 | 0.8793±0.0259 | 0.9065±0.0131 |
| Motif | 0.9476±0.0066 | 0.8750±0.0187 | 0.9032±0.0142 |
| Motif_nobrics | 0.9483±0.0132 | 0.8847±0.0164 | 0.9093±0.0254 |
| Mol+Motif_nobrics | 0.9521±0.0026 | 0.8838±0.0076 | 0.9104±0.0057 |
| Mol+ MACCS | 0.9278±0.0258 | 0.8393±0.0497 | 0.8721±0.0420 |
| Mol+ ERG | 0.9332±0.0292 | 0.8517±0.0614 | 0.8819±0.0506 |
| Mol+ PubChem | 0.9334±0.0284 | 0.8517±0.0597 | 0.8820±0.0492 |
| Mol+ ECFP | 0.9108±0.0367 | 0.8094±0.0649 | 0.8460±0.0570 |
| Mol+FP | 0.9499±0.0156 | 0.8816±0.0391 | 0.9081±0.0303 |
| Motif+FP | 0.9547±0.0098 | 0.8907±0.0187 | 0.9156±0.0254 |
| Mol+Motif+FP | 0.9406±0.0192 | 0.8714±0.0322 | 0.9027±0.0230 |
| Mol+Motif+PK | 0.9536±0.0039 | 0.8884±0.0122 | 0.9139±0.0091 |
| Ours（Mol+Motif) | 0.9575±0.0079 | 0.9048±0.0228 | 0.9251±0.0171 |

The experimental results in Table 7 show that single molecular features (Mol) and motif features (Motif) already possess certain predictive capabilities, but the dual-graph fusion model (Mol+Motif) adopted in this study significantly outperforms any independent feature on all key metrics. In particular, the F1 score reaches 0.9251, powerfully verifying the complementarity of atom-level, molecular-level, and molecular-level local features within the dual-graph architecture, jointly constructing a complete molecular structure representation system.

In the comparative experiment of motif extraction strategies, although the most basic ring-bond extraction strategy (Motif_nobrics) exhibited strong statistical characteristics (F1=0.9093) when used independently, its performance improvement was limited (F1=0.9104) when fused with the molecular graph (Mol+Motif_nobrics). In contrast, although the motif set generated by the BRICS and additional rules strategy (Mol+Motif) adopted in this paper is more complex, it can achieve significant performance improvement (F1=0.9251) after fusion with the molecular graph. This indicates that the motifs extracted by the BRICS and additional rules strategy contain richer functional context and reversible synthetic information, and this deep biological semantics can more effectively supplement atom-level features in the dual-graph fusion architecture, enhancing the model's generalization capability for complex ADR mechanisms.

The comparison with traditional molecular fingerprints reveals the difference between static structural statistics and dynamic functional representation. Fingerprints like PubChem are essentially discrete counts of predefined substructures, and although they showed accuracy close to this model (acc=0.8997) in Table 7, indicating that they can effectively capture basic structural features, their F1 score (0.9183) and Recall are still lower than this model (0.9251). This reflects the limitations of fingerprint methods: fingerprints only characterize "whether a structure exists," but cannot reveal "how the structure acts" through reversible synthetic rules like motifs. PubChem fingerprints rely on fixed substructure matching, and cannot dynamically simulate the in vivo metabolic processes or complex reaction mechanisms of molecules, still possessing "coverage blind spots." Even when combined with molecular graph representations, their essence remains static structural statistics, lacking deep understanding of functional context and biological semantics; predefined fingerprints like MACCS and ErG are difficult to comprehensively capture complex ADR mechanisms due to coverage bias or redundancy, with performance weaker than the way of obtaining motifs via BRICS and additional rules; ECFP fingerprints encode the local chemical environment of atoms through hashing algorithms to generate abstract numerical features, and although they can implicitly capture substructure information, these "substructures" exist in the form of hash values, where key functional substructures are dismantled into multiple isolated substructures, destroying functional synergy. Furthermore, the subgraphs identified by ECFP may lack statistical significance or biological relevance, which presents certain limitations for biological interpretability.

Although the FP strategy improved the performance of single fingerprints through multi-source information complementarity, it still did not surpass the dual-graph architecture of this model, indicating the irreplaceability of rule-based motifs in biological semantic clarity—motifs are function-oriented, while fingerprints are structural-statistics-oriented. When FP is combined with molecular or motif features, although performance improved somewhat, it did not surpass this model, and especially the performance decline of the Mol+Motif+FP combination further indicates that information redundancy exists between traditional fingerprints and motif features, where if a certain motif can completely express a certain biological feature, the fragmented statistical features of fingerprints weaken the integrity of motif functions to a certain extent.

In addition, this study also examined the impact of PK features. Data was first semantically encoded through the BERT pre-trained language model, and then dimensionality reduced via a linear projection layer, thereby obtaining a vectorized representation containing semantic information of pharmacokinetic attributes. However, due to the inherent limitations of PK information in existing databases, pharmacokinetic information for a large number of drugs suffers from serious absence and incompleteness, and this defect in data quality makes it difficult for it to demonstrate its due value in the model. This result inversely validates the robustness and practical value of GM-MLG in performing predictions in cold-start scenarios based on the inherent feature of molecular structure.

In summary, the ablation experiments fully prove the superiority of the drug molecular representation learning method based on the dual-graph architecture proposed in this study. Analysis shows that the static statistical features of traditional fingerprints or sparse pharmacokinetic information both possess significant semantic limitations, while this model successfully constructs a dual-graph representation architecture covering atom-level, molecular-level local (i.e., motif), and molecular-level global hierarchies along with an interactive fusion learning strategy by acquiring fine-grained reversible synthetic motifs via dynamic cutting, effectively solving the cold-start problem caused by the scarcity of multi-dimensional features and high-quality labeled data for new drugs.

4.5.2 Feature Extraction Model Ablation Experiment

To further verify the performance of the feature extraction model, under the premise of maintaining the same input features, training strategies, and comparable parameter scales, this study conducted comparative experiments between the GAT-based feature extraction module and two mainstream GNN models, including GIN [54] and GCN [55]. The values of the three comparative experimental evaluation metrics are detailed in Table 8, where it can be observed that the GAT-based feature extraction model proposed in this study outperforms other GNN models on all performance metrics.

GCN adopts isotropic mean aggregation, and GIN relies only on simple sum aggregation, thus making it difficult to simultaneously model the two types of relationships of molecule-motif and motif-motif, whereas GAT based on the multi-head attention mechanism realizes multi-dimensional feature information interaction between molecules and motifs, dynamically measuring the importance of neighbor nodes by adaptively learning the attention weights of neighbor nodes, thereby focusing on key substructures contributing to ADR prediction in complex molecule and motif interactions; furthermore, the multi-head attention mechanism is able to learn neighborhood representations in parallel in multiple feature subspaces, enabling the model to not only integrate the overall structural information of molecules but also deeply mine the local features of motifs after stacking layers. GAT learns differentiated attention patterns separately for molecule-motif and motif-motif edges, allowing GM-MLG to achieve adaptive modeling for different structural associations, and compared to other GNN models, GM-MLG based on the GAT feature extraction module demonstrated excellent performance on all metrics.。

Table 8 Feature Extraction Model Ablation Experiment

| Method | Precision | Recall | F1 |
| --- | --- | --- | --- |
| GIN | 0.9457±0.0216 | 0.8723±0.0082 | 0.9009±0.0301 |
| GCN | 0.9554±0.0103 | 0.8931±0.0095 | 0.9174±0.0154 |
| Ours | 0.9575±0.0079 | 0.9048±0.0228 | 0.9251±0.0171 |

4.5.3 Prediction Model Ablation Experiment

To further verify the performance of the prediction model, this section replaces the Transformer Decoder-based prediction model in the GM-MLG architecture with Gated Recurrent Unit (GRU), Long Short-Term Memory (LSTM), Bidirectional Long Short-Term Memory (BiLSTM), and LSTM model introducing attention mechanism (LSTM_Att). The experiments in this section aim to deeply explore the impact of the MLG module on the overall performance of the model, evaluating the performance differences of Transformer Decoder, GRU, LSTM, and BiLSTM in the task of ADR prediction, where specific experimental performance metrics are shown in Table 9.

Experimental results indicate that Transformer Decoder significantly outperforms GRU, LSTM, BiLSTM, and LSTM_Att introducing attention mechanism on all metrics. Specifically, GRU and LSTM both adopt gating mechanisms to optimize gradient propagation and capture local temporal features, but are essentially limited by the inherent sequential recursive structure, where the state update at each time step depends on the output of the previous step, leading the model to be extremely prone to gradient vanishing or memory decay when processing long sequences, limiting its ability to model long-distance dependencies of global context. Although BiLSTM makes up for information loss through bidirectional context, and LSTM_Att reinforces key features through attention mechanisms, their underlying encoding methods are still limited by serial logic, making it difficult to completely eliminate bottlenecks in long-distance information transmission. In contrast, Transformer Decoder is based on multi-head self-attention mechanisms and positional encoding, which can simultaneously attend to information at all positions in the sequence and achieve parallel processing, possessing obvious advantages in long-distance dependency modeling and global context capturing. This parallel attention mechanism not only improves generation efficiency but also enhances the model's ability to maintain semantic consistency and coherence during the generation process. Therefore, in generation tasks where ADR prediction needs to model complex long-distance dependency relationships, Transformer Decoder usually demonstrates stronger generation capabilities than traditional LSTM.

Table 9 Prediction Model Ablation Experiment

| Method | Precision | Recall | F1 |
| --- | --- | --- | --- |
| GRU | 0.8992±0.0085 | 0.7565±0.0165 | 0.8095±0.0143 |
| LSTM | 0.9013±0.0145 | 0.8105±0.0249 | 0.8448±0.0231 |
| LSTM_Att | 0.9272±0.0213 | 0.8243±0.0408 | 0.8638±0.0242 |
| BiLSTM | 0.9344±0.0114 | 0.8754±0.0213 | 0.8953±0.0193 |
| Ours | 0.9575±0.0079 | 0.9048±0.0228 | 0.9251±0.0171 |

4.6 Fine-grained Motif Feature Contribution Analysis

This section further designs fine-grained motif feature contribution analysis experiments to quantify the importance of different motif features to adverse reaction prediction results. Specifically, for each encoded motif feature, it is sequentially zeroed out in the input features (i.e., masking that motif feature), and model prediction is performed again under the condition of keeping other features unchanged. By comparing the prediction output under the original complete features with the prediction result after masking the current motif, the difference between the two results is calculated to quantify the contribution degree of that motif. The larger the contribution degree, the stronger the contribution that motif has under that label.

Through the above contribution analysis experiments, this section systematically evaluated and compared key motif features under different adverse reaction labels. As shown in the feature engineering differentiation analysis heatmap in Figure 5, under the label of Nausea (C0027497), motifs with significant contribution degrees include motif 89 and 140, which are 2-Butenylamine and Chlorobenzene functional fragments respectively, suggesting that the potential occurrence mechanism of this adverse reaction may involve the synergistic action of multiple functional groups and fragments. Under the label of Shock (C0036974), although multiple motifs exist in the samples, motif 89 has the highest contribution degree in most samples, showing a dominant structural feature distribution in this adverse reaction. The differential distribution reflects the existence of feature differences in different adverse reactions at the molecular substructure level, suggesting that certain motifs may dominate the occurrence of one or more adverse reactions, while a certain adverse reaction may also be driven by the joint action of multiple motifs, providing important structural information support for the model's feature learning and further mechanism explanation, revealing the non-simple linear additive structure-activity relationship existing between molecular motifs and ADRs, and further providing scientific basis and strategic guidelines for drug "molecular structure optimization design and toxicity ADR regulation".

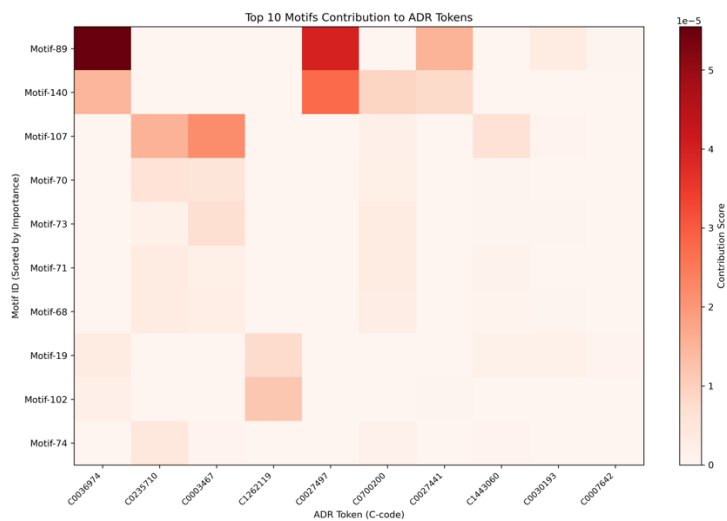

Figure 5 Heatmap of Feature Engineering Differential Analysis

## 5  Conclusion and Future Work

For the first time, this study proposes GM-MLG, an open-ended prediction paradigm for ADRs based on graph-motif feature fusion and multi-label generation, systematically addressing the three core bottlenecks faced by existing ADR prediction methods: cold-start challenges, closed label sets, and inadequate modeling of multi-label dependencies. Targeting the scarcity of multi-dimensional external features in the early stage of new drug development, GM-MLG constructs a dual-graph synergistic representation architecture spanning atomic level, motif local level, and molecular global level, wherein the atomic-level molecular graph characterizes fine-grained topological structures, whereas the Molecule-Motif Association Graph dynamically extracts fine-grained motifs with retrosynthetic attributes based on BRICS and additional rules and explicitly models cross-molecular substructure co-occurrence relationships; this architecture provides sufficient and generalizable structural representation support for the exploration and connectivity of reasoning pathways for ADR prediction in the absence of high-quality labeled data such as protein targets or pharmacokinetics. Furthermore, GM-MLG transforms ADR prediction from traditional multi-label classification into a multi-label generation task based on a Transformer Decoder for the first time, treating the permutation-invariant ADR label set as a discrete token sequence, explicitly modeling long-range dependencies and co-occurrence relationships within the large-scale label space through positional embeddings and Multi-Head Attention mechanisms, and generating prediction results stepwise in an autoregressive manner, thereby dynamically expanding the prediction space to tens of thousands of ADRs under fixed training label scales and limited computational resource overheads, realizing open-ended prediction for potential ADRs of both new and old drugs.

The experimental results fully substantiate the superior performance of GM-MLG. The model significantly surpasses existing SOTA benchmark models in key metrics such as precision, recall, and F1-score, particularly in the evaluation of cold-start scenarios targeting new drugs; experimental results indicate that, although external data such as PK theoretically possess auxiliary value, under the realistic environment where current data are generally scarce and incomplete, such sparse features conversely introduce noise that interferes with prediction and weaken model prediction performance; in contrast, GM-MLG exhibits superior adaptability by virtue of the intrinsic inherent feature of molecular structure. More critically, GM-MLG breaks through the limitations of traditional closed sets and realizes truly open-ended prediction: under the constraint of utilizing only limited labels (200 types) for training, it successfully dynamically expands the prediction space to an open space containing tens of thousands of ADRs, not only capable of performing risk re-evaluation of potential unlabeled side effects for established drugs but also enabling zero-shot inference and prediction for unknown new drugs. On this basis, the validation results of the model on multiple public datasets demonstrate its excellent generalization performance, capable of effectively adapting to prediction tasks under different data distributions. Simultaneously, through the analysis of feature contribution of fine-grained motifs, it reveals the non-simple linear additive structure-activity relationships existing between molecular motifs with retrosynthetic attributes and ADRs, providing an interpretable theoretical basis for drug structural optimization. It is worth noting that, although the model currently takes molecular structure as input, this does not imply the limitation of a single feature, but rather constitutes a multi-scale deep representation system fusing atomic level, motif local level, and molecular global level, and under conditions where more complete and low-noise multi-source data such as pharmacokinetics are obtained in the future, this framework fully supports incorporating them as supplementary features for synergistic modeling. Furthermore, this study has also made pioneering contributions at the data level. Given that the academic community currently lacks public resources directly associating standardized InChI codes with ADR sequences, this study independently constructed and released the benchmark dataset IADRSeq (InChI-ADR-Seq), serving as the first benchmark resource establishing an end-to-end sequence mapping between complete InChI data formats and ADR labels, providing a standardized dataset foundation for validating the potential of generative models in drug safety assessment.

  In summary, GM-MLG reverts to the intrinsic inherent feature of drug molecular structure, and through the deep fusion of dual-graph representation and generative prediction, provides an innovative paradigm with low data dependence, high generalization capability, and open-ended prediction characteristics for drug safety assessment. In the context of the ten-thousand-dimensional label space, this paradigm effectively breaks through the dependence of traditional methods on external prior knowledge, not only providing robust technical support for the prospective risk avoidance of new drugs in cold-start scenarios but also providing a new solution path for the safety re-analysis and systematic discovery of potential adverse reactions of established drugs. Future work will further incorporate multi-dimensional quantitative metrics such as severity grading, occurrence frequency, and clinical priority of adverse reactions, aiming to promote the deeper clinical decision support value of GM-MLG in the context of open-ended prediction.

Acknowledgments:
This study was supported by the Natural Sciences Foundation of Hunan Province (No. 2025JJ50343) and the National Key Research and Development Program of China (No. 2023YFC3503404).

Data and code availability:
The source code and IADRSeq dataset used in this study are publicly available at \url{https://github.com/YuyanPi/GM-MLG}.